\documentclass[twoside]{article}
\usepackage{PRIMEarxiv}

\usepackage{hyperref}
\usepackage{url}
\usepackage[utf8]{inputenc} 
\usepackage[T1]{fontenc}    
\usepackage{booktabs}       
\usepackage{amsfonts}       
\usepackage{nicefrac}       
\usepackage{xcolor}         
\usepackage{balance}
\usepackage{subfigure}
\usepackage{fancyhdr}
\usepackage{graphicx}
\usepackage{bbm}
\usepackage{multicol}
\usepackage{booktabs}
\usepackage{multirow}
\usepackage{threeparttable}
\usepackage{algorithm}  
\usepackage{algorithmicx}  
\usepackage{algpseudocode}  
\usepackage{amsmath} 
\usepackage{color}
\usepackage{float}
\usepackage{natbib}
\usepackage{comment}

\usepackage{rotating}

\usepackage{array}
\newcolumntype{H}{>{\setbox0=\hbox\bgroup}c<{\egroup}@{}}

\usepackage[acronym]{glossaries}
\newacronym{OURMODEL}{PamCGC}{PamC Graph Clustering} 
\newacronym{LOSS}{PamC}{Proxy approximated meta-node Contrastive} 

\title{Efficient block contrastive learning via \\parameter-free meta-node approximation}

\author{Gayan K. Kulatilleke, Marius Portmann \& Shekhar S. Chandra \\
    University of Queensland, Brisbane, Australia.\\
    \small{\texttt{\{g.kulatilleke@uqconnect, marius@itee.uq, shekhar.chandra@uq\}.edu.au}}
}
\begin{document}

\pagestyle{fancy}
\thispagestyle{empty}
\rhead{ \textit{ }} 

\fancyhead[LO]{Kulatilleke et al.}

\maketitle

\begin{abstract}
Contrastive learning has recently achieved remarkable success in many domains including graphs.  
However contrastive loss, especially for graphs, requires a large number of negative samples which is unscalable and computationally prohibitive with a quadratic time complexity.
Sub-sampling is not optimal and incorrect negative sampling leads to sampling bias.
In this work, we propose a meta-node based approximation technique that can (a) proxy \textit{all} negative combinations (b) in quadratic cluster size time complexity, (c) at graph level, not node level, and (d) exploit graph sparsity. 
By replacing node-pairs with additive cluster-pairs, we compute the negatives in cluster-time at graph level.
The resulting \acrfull{LOSS} loss, based on simple optimized GPU operations, captures the full set of negatives, yet is efficient with a linear time complexity.
By avoiding sampling, we effectively eliminate sample bias. We meet the criterion for larger number of samples, thus achieving block-contrastiveness, which is proven to outperform pair-wise losses.
We use learnt soft cluster assignments for the meta-node constriction, and avoid possible heterophily and noise added during edge creation. 
Theoretically, we show that real world graphs easily satisfy conditions necessary for our approximation.
Empirically, we show promising accuracy gains over state-of-the-art graph clustering on 6 benchmarks. Importantly, we gain substantially in efficiency; up to 3x in training time, 1.8x in inference time and over 5x in GPU memory reduction.
Code : https://github.com/gayanku/PAMC
\end{abstract}

\section{Introduction}
Discriminative approaches based on contrastive learning has been outstandingly successful in practice \citep{guo2017improved,wang2020understanding}, achieving state-of-the-art results \citep{chen2020simple} or at times outperforming even supervised learning \citep{logeswaran2018efficient,chen2020simpleVisual}. Specifically in graph clustering, contrastive learning can outperform traditional convolution and attention-based Graph Neural Networks (GNN) on speed and accuracy \citep{kulatilleke2022scgc}.

While traditional objective functions encourage similar nodes to be closer in embedding space, their penalties do not guarantee separation of unrelated graph nodes \citep{zhu2021contrastive}. Differently, many modern graph embedding models \citep{hamilton2017inductive, kulatilleke2022scgc}, use contrastive objectives. These encourage representation of positive pairs to be similar, while making features of the negatives apart in embedding space \citep{wang2020understanding}. A typical deep model consists of a trainable encoder that generates positive and negative node embedding for the contrastive loss \citep{zhu2021contrastive}. It has been shown that convolution is computationally expensive and may not be necessary for representation learning \citep{chen2020simple}. As the requirement for contrastive loss is simply an encoder, recently researchers have been able to produce state-of-the-art results using simpler and more efficient MLP based contrastive loss implementations \citep{hu2021graph, kulatilleke2022scgc}. Thus, there is a rapidly expanding interest and scope for contrastive loss based models.

We consider the following specific but popular \citep{hu2021graph,kulatilleke2022scgc} form of contrastive loss where $\tau$ is the temperature parameter, $\gamma_{ij}$ is the relationship between nodes $i,j$ and the loss for the $i^{th}$ node is: 
\begin{equation}
\ell_{i}=-\log \frac
{\sum_{j=1}^{B} \mathbf{1}_{[j \neq i]}   \gamma_{ij} \cdot \exp \left(\operatorname{sim}\left(\boldsymbol{z}_{i}, \boldsymbol{z}_{j}\right) \cdot \tau\right)}
{\sum_{k=1}^{B} \mathbf{1}_{[k \neq i]} \exp \left(\operatorname{sim}\left(\boldsymbol{z}_{i}, \boldsymbol{z}_{k}\right) \cdot \tau\right)} ,
\label{EQ_TRADITIONAL}
\end{equation}

When no labels are present, sampling of positive and negative nodes plays a crucial role \citep{kipf2016variational} and is a key implementation detail in contrastive methods \citep{velickovic2019deep}. Positive samples in graphs are typically connected by edges \citep{kulatilleke2021fdgatii}, similar to words in a sentence in language modelling \citep{logeswaran2018efficient}. Often data augmentation is used to generate positive samples; \citet{chen2020simpleVisual} used crop, colouring, blurring. However, it is harder to obtain negative samples. 
With no access to labels, negative counterparts are typically obtained via uniform sampling \citep{park2022cgc}, via synthesizing/augmenting \citep{chen2020simpleVisual} or adding noise. Also, in graphs, adjacency information can be exploited to derive negatives \citep{hu2021graph,kulatilleke2022scgc} for feature contrastion. 
However, while graphs particularly suited for contrastive learning, to be effective, a large number of negative samples must be used, along with larger batch sizes and longer training compared to its supervised counterparts \citep{chen2020simpleVisual}.

Unlike other domains, such as vision, negative sample generation brings only limited benefits to graphs \citep{chuang2020debiased,zhu2021contrastive}. 
To understand this phenomenon, observe the raw embedding of USPS image dataset, in the top row of Figure~\ref{fig_embeddings} which looks already clustered. A direct consequence of this is that graphs are more susceptible to sampling bias \citep{chuang2020debiased, zhu2021contrastive}. Thus, graph contrastive learning approaches suffer from insufficient negatives and the complex task of sample generation in addition to $O(N^2)$ time complexity required to contrast every negative node. 

However, what contrastive loss exactly does remain largely a mystery \citep{wang2020understanding}. For example, \citet{arora2019theoretical}'s analysis based on the assumption of latent classes provides good theoretical insights, yet their explanation on representation quality dropping with large number of negatives is inconsistent with experimental findings \citep{chen2020simpleVisual}. Contrastive loss is seen as maximizing mutual information (MI) between two views. Yet, contradictorily, tighter bound on MI can lead to poor representations \citep{wang2020understanding}.

\textbf{Motivation}:
\citet{wang2020understanding} identifies alignment and uniformity as key properties of contrastive loss: alignment encourages encoders to assign similar features to similar samples; uniformity encourages a feature distribution that preserves maximal information. It is fair to assume that latent clusters are dissimilar. Even with the rare possibility of two identical cluster centers initially, one will usually change or drift apart. It is intuitive that cluster centers should be uniformly distributed in the hyperspace, similar to nodes, in order to preserve as much information of the data as possible. Uniformly distributing points on a hyperspace is defined as minimizing the total pairwise potential w.r.t. a certain kernel function and is well-studied \citep{wang2020understanding}.

Thus, we are naturally motivated to use the cluster centers as meta-nodes for negative contrastion. By aggregation, all its constituent nodes cab be affected. Thus, we avoid sampling, effectively eliminate sample bias, and also meet the criterion of larger number of samples. Learned soft cluster assignments can avoid possible heterophily and add robustness to noise in edge construction. 

In this work, we propose a novel contrastive loss, \acrshort{LOSS}, which uses paramaterless proxy meta-nods to approximate negative samples. Our approach indirectly uses the full set of negative samples and yet is efficient with a time complexity of $O(N)$. Not only does \acrshort{OURMODEL}, based on \acrshort{LOSS}, outperform or match previous work, but it is also simpler than any prior negative sample generation approach, faster and uses relatively less GPU memory. It can be incorporated to any contrastive learning-based clustering model with minimal modifications, and works with diverse data, as we demonstrate using benchmark datasets from image, text and graph modalities.

To summarize, our main contributions are:
\begin{itemize}
\item We introduce an efficient novel parameter-free proxy, \acrshort{LOSS}, for negative sample approximation that is  scalable, computationally efficient and able to include \textit{all} samples. It works with diverse data, \textit{including} graphs.
We claim \acrshort{LOSS} is the first to \textit{implicitly} use the \textit{whole} graph with $O(N)$ time complexity, in addition to further 3-fold gains.
\item We provide theoretical proof and show that real world graphs always satisfies the necessary conditions, and that \acrshort{OURMODEL} is block-contrastive, known to outperform pair-wise losses.
\item Extensive experiments on 6 benchmark datasets show \acrshort{OURMODEL}, using proposed \acrshort{LOSS}, is on par with or better than  state-of-the-art graph clustering methods in accuracy while achieving 3x training time, 1.8 inference time and 5x GPU memory efficiency.
\end{itemize}

\section{Implementation}\label{ProposedModel}
\begin{figure}[t]
  \centering
  \includegraphics[width=1.0\textwidth]{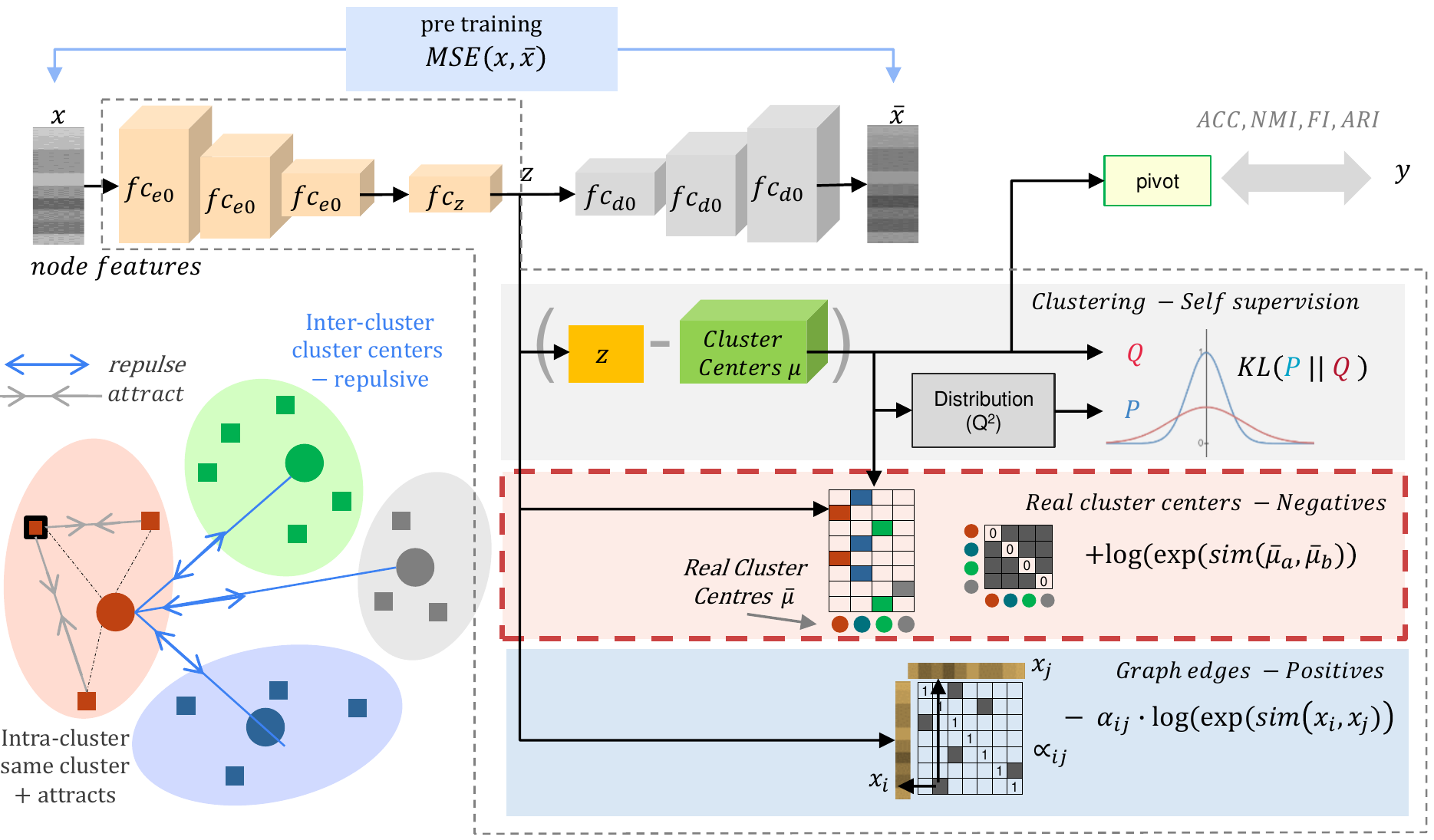}
  \caption{\acrshort{OURMODEL} jointly learns structure and clustering via probabilistic soft assignment which is used to derive the real cluster centers $\hat{\mu}$, used as proxy for negative samples. Grey dotted section outlines the training components. Cluster centroids $\mu$ are obtained by pre-training an AE for reconstruction. Red dotted section is our core contribution: we use $\hat{\mu}$ as an efficient approximation, computing centroid-pairs instead of node-pairs, achieve block-contrastivness and do so at graph level, not instance level.}
  \label{fig_model}
\end{figure}

First we describe \acrshort{LOSS}, which is our parameter-free proxy to efficiently approximate the negative samples, as shown in Figure~\ref{fig_model}. Next, we introduce \acrshort{OURMODEL}, a self-supervised model based on \acrshort{LOSS} to simultaneously learn discriminative embeddings and clusters. 

\subsection{Negative sample approximation by meta-node proxies} \label{Approximation}
Contrastive loss makes positive or connected nodes closer and negative or unconnected nodes further away in the feature space \citep{kulatilleke2022scgc}. However, in order to be effective, \textit{all} negative nodes need to be contrasted with $x_i$ which is computationally expensive. A cluster centre is formed by combining all member nodes, and can be seen as an aggregated representation, or a proxy, of its compositional elements. Motivated by this, we use the cluster centres to enforce negative contrastion. Specifically, we contrast every cluster centre $\hat{\mu_i}$ with every cluster centre $\hat{\mu_j}$ where $i \ne j$.

Following \citet{arora2019theoretical, chuang2020debiased}, we assume an underlying set of discrete latent classes $C$ which represents semantic content,
i.e., similar nodes $x_i, x_j$ are in the same latent class $\hat{\mu}$. Thus, we derive our proxy for negative samples as:
\begin{equation}
\ell_{proxy}=\log{
\sum_{a=1}^{C} \sum_{b=1}^{C} \mathbf{1}_{[a \neq b]} 
\exp \left(\operatorname{sim}\left(\boldsymbol{\hat{\mu}}_{a}, \boldsymbol{\hat{\mu}}_{b}\right) \cdot \tau\right)
},
\label{EQ_PROXY}
\end{equation}
Note that, $\ell_{proxy}$ contains no $i$ or $j$ terms! resulting in \textit{three} fold gains. Firstly, we replace $\sum_{i=1}^{N}$, with a more efficient $\sum_{a=1}^{C}$ where $N \gg C$, typically many magnitudes, in almost all datasets, as evident from Table~\ref{TABLE_datasets}. Secondly, the $\ell_{proxy}$ is at \textit{graph level} with time complexity of $O(N)$ rather than an instance level $O(N^2)$. Finally, given real world graphs (especially larger graphs,) are sparse, a sparse implantation for the positives, using edge-lists, will result in a $third$ efficiency gain, which is only possible by not having to operate on the negatives explicitly. 

Note that a prerequisite of the proxy approximation is the availability of labels to construct the learned cluster centers $\hat{\mu}$, which we explain in the next section. Thus, the complete graph level contrastive loss can be expressed as:
\begin{equation}
\ell_{Pcontrast}=- \frac{1}{N} \sum_{i=1}^{N} \log \sum_{j=1}^{N} \mathbf{1}_{[j \neq i]}   \gamma_{ij}  \exp \left(\operatorname{sim}\left(\boldsymbol{z}_{i}, \boldsymbol{z}_{j}\right) \cdot \tau\right) \\
+ \ell_{proxy},
\label{EQ_PROXYFULL}
\end{equation}

\textbf{Theoretical explanation.} 
The standard contrastive loss uses Jensen-Shannon divergence, which yields $\log 2$ constant and vanishing gradients for disjoint distributions of positive and negative sampled pairs \citep{zhu2021contrastive}. However, in the proposed method, positive pairs are necessarily edge-linked (either explicitly or via influence \citep{kulatilleke2022scgc}), and unlikely to be disjoint. Using meta-nodes for negatives, which are compositions of multiple nodes, lowers the possibility of disjointness. An algorithm using the average of the positive and negative samples in blocks as a proxy instead of just one point has a strictly better bound due to Jensen’s inequality getting tighter and is superior compared to their equivalent of element-wise contrastive counterparts \citep{arora2019theoretical}.
The computational and time cost is a direct consequence of node level contrastion. Given, $N\gg clusters$, we circumvent the problem of large $N$ by proposing a proxy-ed negative contrastive objective that operates directly at the cluster level.

\underline{Establishing mathematical guarantee:}
Assume node embeddings $Z=\{z_1,z_2,z_3 \dots z_N\}$, clusters $\mu=\{\mu_1,\mu_2 \dots \mu_C\}$, a label assignment operator $\operatorname{label}(z_i)$ such that $\mu_a = \sum_{i=1}^{N} \mathbf{1}_{[i \in \operatorname{label}(z_i)=a]}\cdot z_i$, a temperature hyperparameter $\tau$ and,
\begin{equation}
     \operatorname{similarity}(i, j, z_i,z_j) = \operatorname{sim}(z_i,z_j) \begin{cases}
        0,& \text{$i=j$} \\
        \frac{z_i \cdot z_j}{\|z_i\| \|z_j\|} ,& \text{$i \ne j$}\\
\end{cases}
\label{EQ_SIM}
\end{equation}
Using $\operatorname{sim}(z_i,z_j)$ as the shorthand notation for $\operatorname{similarity}(i, j, z_i,z_j)$, the classic contrastive loss is:
\begin{equation}
loss_{NN} = \frac{1}{N}  \sum _{i=1}^{N} \log\left[ 
    \sum _{j=1}^{N} \exp ( \operatorname{sim}(i,j,z_i,z_j)\tau)
\right],
\label{EQ_LOSSNN}
\end{equation}

Similarly, we can express the cluster based contrastive loss as:
\begin{equation}
loss_{CC} = \frac{1}{C}  \sum _{a=1}^{C} \log\left[ 
    \sum _{b=1}^{M} \exp ( \operatorname{sim}(a,b,\mu_a,\mu_b)\tau)
\right]
\label{EQ_LOSSCC}
\end{equation}

As $ 0 \leq \operatorname{sim} \leq 1.0$, we establish the condition for our inequality as;
\begin{align}
  \frac{loss_{NN}}{loss_{CC}} >  \frac{\log(N)}{\log \left[ 1 + (C-1)e^{\tau}\right]}
\label{EQ_INE}											  
\end{align}
We provide the full derivation in Appendix~\ref{Appendix_PROOF1}.

\begin{figure}[th]
  \centering
  \includegraphics[width=0.40\columnwidth]{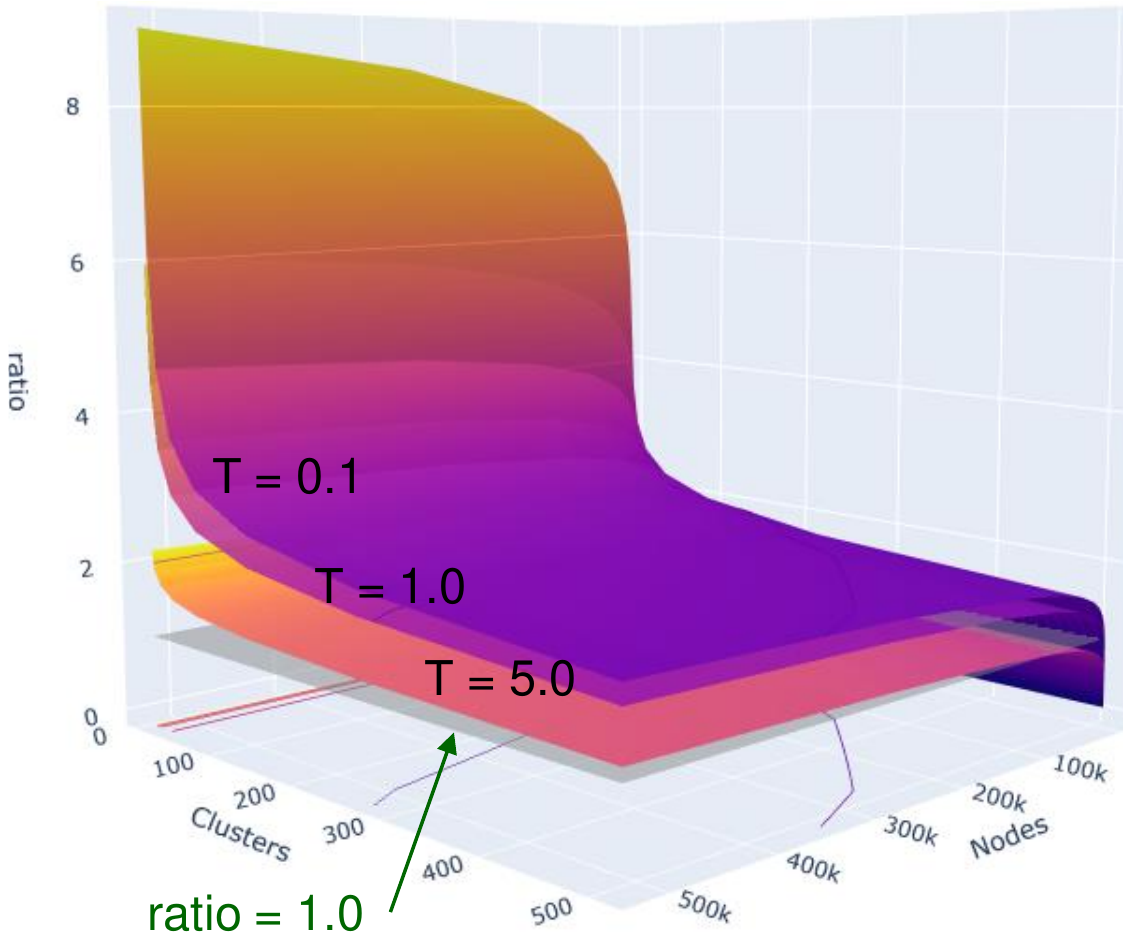}
  \caption{Nodes $N$ vs Clusters $C$ with different $\tau$ temperature values. Grey surface shows the $ratio=1.0$ inequality boundary. Generally, real world graphs satisfy the condition $ratio>1.0$ easily. Best viewed in color.}
  \label{fig_NCMap1}
\end{figure}
As $C > 1$ (minimum 2 are needed for a cluster), and $log(x):x>0$ is strictly increasing, $N > 1 + (C-1)e^{\tau} $ is the necessary condition, which is easily satisfied for nearly all real world datasets and as seen in Figure~\ref{fig_NCMap1} for multiple $\tau$ temperatures.

Thus, as $loss_{NN} > loss_{CC}$,  $loss_{NN}$ upper bounds $loss_{CC}$, the more efficient variant. Additionally $loss_{CC}$ benefits from block-contrastiveness \citep{arora2019theoretical}, achieves a lower minima and uses the \textit{fullest possible} negative information.
We also show, experimentally, that minimizing $loss_{CC}$ results in effective, and sometimes better, representations for downstream tasks. 


\subsection{Constructing the meta-node cluster centres ($\hat{\mu}$) }
In order to derive the real cluster centers $\hat{\mu}$, which is distinct from the learnt cluster centers $\mu$, we simply aggregate all the node embedding $z$ of a cluster using its $\operatorname{label}$. Even with unlabeled data, $\operatorname{label()}$ can be accomplished using predicted soft labels. The intuition here is that, during back-propagation, the optimization process will update the constituent node embeddings, $z$, to incorporate negative distancing. Thus,
\begin{equation}
    \hat{\mu_c}=\frac{1}{N}
    \sum_{i=1}^{N} \mathbf{1}_{[i \in \operatorname{label}(c)]} z_i,
    \label{EQ_MUHAT}
\end{equation}
where $\operatorname{label(c)}$ is either the ground-truth or learnt soft labels. Accordingly, our proxy can equally be used in supervised and unsupervised scenarios and has a wider general applicability as an improvement of the contrastive loss at large. Finlay, Equation~\ref{EQ_MUHAT} can be realized with $softmax()$ and $mean()$ operations, which are well optimized GPU primitives in any machine learning framework. We provide a reference pytorch implementation.

\subsection{Obtaining the soft labels}
Graph clustering is essentially unsupervised. To this end, following \citet{xie2016unsupervised,guo2017improved,wang2019attributed,kulatilleke2022scgc}, we use probability distribution derived soft-labels and a self-supervision mechanism for cluster enhancement. Specifically, we obtain soft cluster assignments probabilities $q_{iu}$ for embedding $z_i$ and cluster centre $\mu_u$. In order to handle differently scaled clusters and be computationally convenient \citep{wang2019attributed}, we use the student's $t$-distribution \citep{maaten2008visualizing} as a kernel for the similarity measurement between the embedding and centroid:
\begin{equation} 
q_{iu} = \frac{(1+\left\|z_i - {\mu}_u\right\|^2/\eta)^{-\frac{\eta+1}{2}}}
        {\sum_{u'}(1+\left\|z_i - {\mu}_{u'}\right\|^{2}/\eta)^{-\frac{\eta+1}{2}}},
\label{eq-QIJ}
\end{equation} 
where, $\eta$ is the Student’s $t$-distribution's degree of freedom. Cluster centers $\mu$ are initialized by $K$-means on embeddings from the pre-trained AE. We use $Q=[q_{iu}]$ as the distribution of the cluster assignments of all samples, and $\eta$=1 for all experiments following \citet{bo2020structural,peng2021attention,kulatilleke2022scgc}

Nodes closer in embedding space to a cluster center has higher soft assignment probabilities in $Q$. A target distribution $P$ that emphasizes the \textit{confident} assignments is obtained by squaring and normalizing $Q$, given by : 
\begin{equation} 
p_{iu} = \frac{q_{iu}^2/\sum_i q_{iu}}{\sum_k{(q_{ik}^2/\sum_i q_{ik})}},
\label{eq-PIJ}
\end{equation} 
where $\sum_{i}q_{iu}$ is the soft cluster frequency of centroid $u$.

Following \citet{kulatilleke2022scgc}, we minimize the KL divergence between $Q$ and $P$ distributions, which forces the current distribution $Q$ to approach the more confident target distribution $P$. KL divergence updates models more gently and lessens severe disturbances on the embeddings \citep{bo2020structural}. Further, it can accommodate both the structural and feature optimization targets of \acrshort{OURMODEL}. We self-supervise cluster assignments \footnote{We follow \citet{bo2020structural} use of the term 'self-supervised' to be consistent with the GCN training method.} by using distribution $Q$ to target distribution $P$, which then supervises the distribution $Q$ in turn by minimizing the KL divergence as:  
\begin{equation} 
loss_{cluster}= KL(P||Q)=\sum_{i}\sum_{u}p_{iu}log\frac{p_{iu}}{q_{iu}},
\label{eq-KLPQ}    
\end{equation} 

\textbf{The final proposed model,}
after incorporating \acrshort{LOSS} contrastive objective with self-supervised clustering, where $\alpha>0$ controls structure incorporation and $\beta>0$ controls cluster optimization is: 
\begin{equation} 
    \rm{\acrshort{OURMODEL}:} \quad L_{final} =  \alpha \ell_{Pcontrast}(K,\tau) + \beta  loss_{cluster} ,
\label{eq-FINALLOSS}
\end{equation}

\subsection{Complexity Analysis} \label{Complexity}
Given input data dimension $d$ and AE layer dimensions of $d_{1}, d_{2}, \cdots, d_{L}$, following \citet{kulatilleke2022scgc},  $O_{AE}=\mathcal{O}(Nd^{2}d^{2}_{1}...d^{2}_{L/2})$ for \acrshort{OURMODEL}-AE. Assuming $K$ clusters, from Equation~\ref{eq-QIJ}, the time complexity is $O_{cluster} = \mathcal{O}(NK+N\log N)$ following \citet{xie2016unsupervised}.

For \acrshort{LOSS}, we only compute $\|z\|_{2}^{2}$ and $z_{i}\cdot z_{j}$ for the actual positive edges $E$ using sparse matrix resulting in a time complexity $O_{+} = \mathcal{O}(NEd_z)$,  linear with the number of edges $E$, with $d_z$ embedding dimension. For the negatives, we use the meta-node based negatives $O_{-} = \mathcal{O}(CC)$ where $C$ is the meta-node. Note that, for real graphs, $N \gg C$ in many magnitudes. Thus, the overall time complexity is linearly related to the number of samples and edges.

\section{Experiments}\label{Experiments}
We evaluate \acrshort{OURMODEL} on transductive node clustering comparing to state-of-the-art self-supervised, contrastive and (semi-)supervised methods. 

\begin{table}
\caption{Statistics of the datasets  (left) and \acrshort{OURMODEL} hyperparameters (right).}
\label{TABLE_datasets}
\centering
\begin{tabular}{lcrrrrrrrrr}
\cmidrule(r){1-5} \cmidrule(r){6-11} 
Dataset  & Type   & \textbf{N}odes & \textbf{C}lasses & \textbf{d}imension & & $\alpha$ & $\beta$ & K  & $\tau$ & LR \\ \cmidrule(r){1-5} \cmidrule(r){6-11} 
USPS     & Image  & 9298    & 10      & 256      & & 2  & 2 & 4 & 0.5  &  $10^{-3}$ \\
HHAR     & Record & 10299   & 6       & 561      & & 0.5  & 12.5 & 2 & 1.5  &  $10^{-3}$ \\
REUT  & Text   & 10000   & 4       & 2000     & & 1  & 0.2 & 1 & 0.25  &  $10^{-4}$ \\
ACM      & Graph  & 3025    & 3       & 1870     & & 0.5  & 0.5 & 1 & 0.5  &  $10^{-3}$ \\
CITE & Graph  & 3327    & 6       & 3703     & & 2  & 2 & 1 & 1  &  $10^{-3}$ \\ 
DBLP     & Graph  & 4057    & 4       & 334      & & 2  & 2.5 & 3 & 0.5  &  $10^{-3}$ \\ \bottomrule
\end{tabular}
\end{table}

\textbf{Datasets.} 
\label{dataset}Following \citet{bo2020structural,peng2021attention,kulatilleke2022scgc}, experiments are conducted on six common clustering benchmarks, which includes one image dataset (USPS \citep{le1990handwritten}), one sensor data dataset (HHAR \citep{stisen2015smart}), one text dataset (REUT \citep{lewis2004rcv1}) and three citation graphs (ACM\footnotemark[2], CITE\footnotemark[4], and  DBLP\footnotemark[3]). For the non-graph data, we use undirected $\emph{k}$-nearest neighbour (KNN \citep{altman1992introduction}) to generate adjacency matrix $\mathbf{A}$ following \citet{bo2020structural, peng2021attention}. Table~\ref{TABLE_datasets} summarizes the datasets.
\footnotetext[2]{http://dl.acm.org/}
\footnotetext[3]{https://dblp.uni-trier.de}
\footnotetext[4]{http://citeseerx.ist.psu.edu/index}

\textbf{Baseline Methods.}
We compare with multiple models.
K-means \citep{hartigan1979algorithm} is a classical clustering method using raw data.
AE \citep{hinton2006reducing} applies K-means to deep representations learned by an auto-encoder. 
DEC \citep{xie2016unsupervised} clusters data in a jointly optimized feature space.
IDEC \citep{guo2017improved} enhances DEC by adding KL divergence-based reconstruction loss.
Following models exploit graph structures during clustering:
SVD~\citep{golub1971singular} applies singular value decomposition to the adjacency matrix.
DGI~\citep{velickovic2019deep} learns embeddings by maximizing node MI with the graph.
GAE \citep{kipf2016variational} combines convolution with the AE.
ARGA \citep{pan2018adversarially} uses an adversarial regularizer to guide the embeddings learning. 
Following deep graph clustering jointly optimize embeddings and graph clustering:
DAEGC \citep{wang2019attributed}, uses an attentional neighbor-wise strategy and clustering loss.  
SDCN \citep{bo2020structural}, couples DEC and GCN via a fixed delivery operator and uses feature reconstruction.
AGCN \citep{peng2021attention}, extends SDCN by adding an attention-based delivery operator and uses multi scale information for cluster prediction.
CGC \citep{park2022cgc} uses a multi-level, hierarchy based contrastive loss.
SCGC and SCGC* \citep{kulatilleke2022scgc} uses block contrastive loss with an AE and MLP respectively.
The only difference between SCGC* and \acrshort{OURMODEL} is the novel \acrshort{LOSS} loss, Also as SCGC* is the current state-of-the-art. Thus, it is used as the benchmark.

\textbf{Evaluation Metrics.}
Following \citet{bo2020structural,peng2021attention}, we use Accuracy (ACC), Normalized Mutual Information (NMI), Average Rand Index (ARI), and macro F1-score (F1) for evaluation. For each, larger values imply better clustering.

\subsection{Implementation} \label{Implementation}
The positive component of our loss only requires the actual connections and can be efficiently represented by sparse matrices. Further, the negative component of the loss is graph-based, and not instance based, thus needs to be computed only once per epoch. Thus, by decoupling the negatives, our loss is inherently capable of batching and is trivially parallelizable. Computation of the negative proxy, which is only $C \cdot C$ does not even require a GPU!

For fair comparison, we use the same $500-500-2000-10$ AE dimensions as in \citet{guo2017improved,bo2020structural,peng2021attention,kulatilleke2022scgc} and the same pre-training procedure, i.e. $30$ epochs; learning rate of $10^{-3}$ for USPS, HHAR, ACM, DBLP and $10^{-4}$ for REUT and CITE; batch size of $256$. We made use of the publicly available pre-trained AE from \citet{bo2020structural}. 
We use a once computed edge-list for training, which is not needed during inference. For training, for each dataset, we initialize the cluster centers from $K$-means and repeat the experiments 10 times with $200$ epochs to prevent extreme cases. We cite published accuracy results from \citet{bo2020structural, peng2021attention, kulatilleke2022scgc} for other models.

For all timing and memory experiments, we replicate the exact same training loops, including internal evaluation metric calls, when measuring performance for fair comparison. Our code will be made publicly available.

\begin{table*}
\scriptsize
\caption{Clustering performance the three graph datasets (mean$\pm$std). Best results are \textbf{bold}. Results reproduced from \citet{bo2020structural, peng2021attention, kulatilleke2022scgc, park2022cgc}. SCGC \citep{kulatilleke2022scgc} uses neighbor based contrastive loss with AE while SCGC* variant uses $r$-hop cumulative Influence contrastive loss with MLP, same as our \acrshort{OURMODEL}}.
\label{table_graph_results}
\centering
\setlength{\tabcolsep}{0.7mm}{
\begin{tabular}{lrrrrrrrrrrrr}
\toprule
\multirow{2}{*}{Method} & \multicolumn{4}{c}{DBLP} & \multicolumn{4}{c}{ACM} & \multicolumn{4}{c}{CITE}  \\ \cmidrule(r){2-5} \cmidrule(r){6-9} \cmidrule(r){10-13} 
         & ACC & NMI & ARI & F1 & ACC & NMI & ARI & F1 & ACC & NMI & ARI & F1\\ \midrule
K-means        & 38.7$\pm$0.7          & 11.5$\pm$0.4          & 7.0$\pm$0.4           & 31.9$\pm$0.3          & 67.3$\pm$0.7          & 32.4$\pm$0.5          & 30.6$\pm$0.7          & 67.6$\pm$0.7          & 39.3$\pm$3.2          & 16.9$\pm$3.2          & 13.4$\pm$3.0          & 36.1$\pm$3.5          \\
AE              & 51.4$\pm$0.4          & 25.4$\pm$0.2          & 12.2$\pm$0.4          & 52.5$\pm$0.4          & 81.8$\pm$0.1          & 49.3$\pm$0.2          & 54.6$\pm$0.2          & 82.0$\pm$0.1          & 57.1$\pm$0.1          & 27.6$\pm$0.1          & 29.3$\pm$0.1          & 53.8$\pm$0.1          \\
DEC           	& 58.2$\pm$0.6          & 29.5$\pm$0.3          & 23.9$\pm$0.4          & 59.4$\pm$0.5          & 84.3$\pm$0.8          & 54.5$\pm$1.5          & 60.6$\pm$1.9          & 84.5$\pm$0.7          & 55.9$\pm$0.2          & 28.3$\pm$0.3          & 28.1$\pm$0.4          & 52.6$\pm$0.2          \\
IDEC       			& 60.3$\pm$0.6          & 31.2$\pm$0.5          & 25.4$\pm$0.6          & 61.3$\pm$0.6          & 85.1$\pm$0.5          & 56.6$\pm$1.2          & 62.2$\pm$1.5          & 85.1$\pm$0.5          & 60.5$\pm$1.4          & 27.2$\pm$2.4          & 25.7$\pm$2.7          & 61.6$\pm$1.4          \\
SVD              & 29.3$\pm$0.4          & 0.1$\pm$0.0           & 0.0$\pm$0.1           & 13.3$\pm$2.2          & 39.9$\pm$5.8          & 3.8$\pm$4.3           & 3.1$\pm$4.2           & 30.1$\pm$8.2          & 24.1$\pm$1.2          & 5.7$\pm$1.5           & 0.1$\pm$0.3           & 11.4$\pm$1.7          \\
DGI				& 32.5$\pm$2.4          & 3.7$\pm$1.8           & 1.7$\pm$0.9           & 29.3$\pm$3.3          & 88.0$\pm$1.1          & 63.0$\pm$1.9          & 67.7$\pm$2.5          & 88.0$\pm$1.0          & 64.1$\pm$1.3          & 38.8$\pm$1.2          & 38.1$\pm$1.9          & 60.4$\pm$0.9          \\
GAE				& 61.2$\pm$1.2          & 30.8$\pm$0.9          & 22.0$\pm$1.4          & 61.4$\pm$2.2          & 84.5$\pm$1.4          & 55.4$\pm$1.9          & 59.5$\pm$3.1          & 84.7$\pm$1.3          & 61.4$\pm$0.8          & 34.6$\pm$0.7          & 33.6$\pm$1.2          & 57.4$\pm$0.8          \\
VGAE			& 58.6$\pm$0.1          & 26.9$\pm$0.1          & 17.9$\pm$0.1          & 58.7$\pm$0.1          & 84.1$\pm$0.2          & 53.2$\pm$0.5          & 57.7$\pm$0.7          & 84.2$\pm$0.2          & 61.0$\pm$0.4          & 32.7$\pm$0.3          & 33.1$\pm$0.5          & 57.7$\pm$0.5          \\
ARGA            & 61.6$\pm$1.0          & 26.8$\pm$1.0          & 22.7$\pm$0.3          & 61.8$\pm$0.9          & 86.1$\pm$1.2          & 55.7$\pm$1.4          & 62.9$\pm$2.1          & 86.1$\pm$1.2          & 56.9$\pm$0.7          & 34.5$\pm$0.8          & 33.4$\pm$1.5          & 54.8$\pm$0.8          \\
DAEGC  			& 62.1$\pm$0.5          & 32.5$\pm$0.5          & 21.0$\pm$0.5          & 61.8$\pm$0.7          & 86.9$\pm$2.8          & 56.2$\pm$4.2          & 59.4$\pm$3.9          & 87.1$\pm$2.8          & 64.5$\pm$1.4          & 36.4$\pm$0.9          & 37.8$\pm$1.2          & 62.2$\pm$1.3          \\
CGC				& 77.6$\pm$0.5 			& 46.1$\pm$0.6 			& 49.7$\pm$1.1 			& 77.2$\pm$0.4 			& 92.3$\pm$0.3 			& 72.9$\pm$0.7 			& 78.4$\pm$0.6 			& 92.3$\pm$0.3 			& 69.6$\pm$0.6 			& 44.6$\pm$0.6 			& 46.0$\pm$0.6 			& 65.5$\pm$0.7 			\\ \hline

SDCN     		& 68.1$\pm$1.8          & 39.5$\pm$1.3          & 39.2$\pm$2.0          & 67.7$\pm$1.5          & 90.5$\pm$0.2          & 68.3$\pm$0.3          & 73.9$\pm$0.4          & 90.4$\pm$0.2          & 66.0$\pm$0.3          & 38.7$\pm$0.3          & 40.2$\pm$0.4          & 63.6$\pm$0.2	     	\\ 
AGCN      		& 73.3$\pm$0.4     		& 39.7$\pm$0.4     		& 42.5$\pm$0.3     		& 72.8$\pm$0.6     		& 90.6$\pm$0.2     		& 68.4$\pm$0.5     		& 74.2$\pm$0.4     		& 90.6$\pm$0.2     		& 68.8$\pm$0.2     		& 41.5$\pm$0.3     		& 43.8$\pm$0.3     		& 62.4$\pm$0.2          \\ 
SCGC			& \rm{77.7$\pm$0.1} 	& \rm{47.1$\pm$0.2} 	& \rm{51.2$\pm$0.2} 	& \rm{77.3$\pm$0.1} 	& \rm{92.6$\pm$0.0} 	& \rm{73.3$\pm$0.0} 	& \rm{79.2$\pm$0.0}  	& \rm{92.5$\pm$0.0} 	& \rm{73.2$\pm$0.1} 	& \rm{46.8$\pm$0.1} 	& \rm{50.0$\pm$0.1} 	& \rm{63.3$\pm$0.0} 	\\
SCGC*			& \rm{77.7$\pm$0.1} 	& \rm{47.1$\pm$0.1} 	& \rm{50.2$\pm$0.1} 	& \rm{77.5$\pm$0.1} 	& \textbf{92.6$\pm$0.0} 	& \textbf{73.7$\pm$0.1} 	& \textbf{79.4$\pm$0.1}  	& \textbf{92.6$\pm$0.0} 	& \rm{73.3$\pm$0.0} 	& \rm{46.9$\pm$0.0} 	& \textbf{50.2$\pm$0.0} 	& \textbf{63.4$\pm$0.0} 	\\ \hline

\acrshort{OURMODEL} 						& \textbf{79.6$\pm$0.0} 	& \textbf{49.2$\pm$0.1} 	& \textbf{54.7$\pm$0.1} 	& \textbf{79.0$\pm$0.1} 	& \rm{92.5$\pm$0.0} 	& \rm{73.7$\pm$0.1} 	& \rm{79.2$\pm$0.1}  	& \rm{92.5$\pm$0.0} 	& \textbf{73.3$\pm$0.2} 	& \textbf{47.3$\pm$0.3} 	& \rm{50.1$\pm$0.4} 	& \textbf{63.4$\pm$0.2} 	\\

\bottomrule
\end{tabular}
}
\end{table*}

\begin{table*}
\scriptsize
\caption{Clustering performance the three non-graph datasets (mean$\pm$std). Best results are \textbf{bold}; second best is \underline{underlined}. Results reproduced from \citet{bo2020structural, peng2021attention, kulatilleke2022scgc}. SCGC \citep{kulatilleke2022scgc} uses neighbour based contrastive loss with AE while SCGC* variant uses $r$-hop cumulative Influence contrastive loss with MLP, same as our \acrshort{OURMODEL}}.
\label{table_non_graph_results}
\centering
\setlength{\tabcolsep}{1.1mm}{
\begin{tabular}{llcHHHcccccccc} 
\hline
Dataset                   & Metric & $K$-means      & AE             & DEC            & IDEC           & GAE            & VGAE           & DAEGC          & SDCN                    & AGCN                 &  SCGC    			  &  SCGC*  				 & \acrshort{OURMODEL}\\ \hline
\multirow{4}{*}{USPS}     & ACC    & 66.82$\pm$0.04 & 71.04$\pm$0.03 & 73.31$\pm$0.17 & 76.22$\pm$0.12 & 63.10$\pm$0.33 & 56.19$\pm$0.72 & 73.55$\pm$0.40 & 78.08$\pm$0.19          & \rm{80.98$\pm$0.28}  &  82.90$\pm$0.08      &  \textbf{84.91$\pm$0.06} &  \underline{84.20$\pm$0.24} \\
                          & NMI    & 62.63$\pm$0.05 & 67.53$\pm$0.03 & 70.58$\pm$0.25 & 75.56$\pm$0.06 & 60.69$\pm$0.58 & 51.08$\pm$0.37 & 71.12$\pm$0.24 & 79.51$\pm$0.27          & \rm{79.64$\pm$0.32}  &  82.51$\pm$0.07      &  \textbf{84.16$\pm$0.10} &  \underline{80.32$\pm$0.38} \\
                          & ARI    & 54.55$\pm$0.06 & 58.83$\pm$0.05 & 63.70$\pm$0.27 & 67.86$\pm$0.12 & 50.30$\pm$0.55 & 40.96$\pm$0.59 & 63.33$\pm$0.34 & 71.84$\pm$0.24          & \rm{73.61$\pm$0.43}  &  76.48$\pm$0.11      &  \textbf{79.50$\pm$0.06} &  \underline{77.75$\pm$0.56} \\
                          & F1     & 64.78$\pm$0.03 & 69.74$\pm$0.03 & 71.82$\pm$0.21 & 74.63$\pm$0.10 & 61.84$\pm$0.43 & 53.63$\pm$1.05 & 72.45$\pm$0.49 & 76.98$\pm$0.18          & \rm{77.61$\pm$0.38}  &  80.06$\pm$0.05      &  \textbf{81.54$\pm$0.06} &  \underline{78.82$\pm$0.17} \\ \hline
						  
\multirow{4}{*}{HHAR}     & ACC    & 59.98$\pm$0.02 & 68.69$\pm$0.31 & 69.39$\pm$0.25 & 71.05$\pm$0.36 & 62.33$\pm$1.01 & 71.30$\pm$0.36 & 76.51$\pm$2.19 & 84.26$\pm$0.17          & \rm{88.11$\pm$0.43}  &  \textbf{89.49$\pm$0.22}     & \textbf{89.36$\pm$0.16}  &  84.94$\pm$1.09\\
                          & NMI    & 58.86$\pm$0.01 & 71.42$\pm$0.97 & 72.91$\pm$0.39 & 74.19$\pm$0.39 & 55.06$\pm$1.39 & 62.95$\pm$0.36 & 69.10$\pm$2.28 & 79.90$\pm$0.09          & \underline{82.44$\pm$0.62}  &  84.24$\pm$0.29      &  \textbf{84.50$\pm$0.41} &  79.54$\pm$0.65\\
                          & ARI    & 46.09$\pm$0.02 & 60.36$\pm$0.88 & 61.25$\pm$0.51 & 62.83$\pm$0.45 & 42.63$\pm$1.63 & 51.47$\pm$0.73 & 60.38$\pm$2.15 & 72.84$\pm$0.09          & \underline{77.07$\pm$0.66}  &  \textbf{79.28$\pm$0.28}     & \textbf{79.11$\pm$0.18}  &  72.57$\pm$1.20\\
                          & F1     & 58.33$\pm$0.03 & 66.36$\pm$0.34 & 67.29$\pm$0.29 & 68.63$\pm$0.33 & 62.64$\pm$0.97 & 71.55$\pm$0.29 & 76.89$\pm$2.18 & 82.58$\pm$0.08          & \underline{88.00$\pm$0.53}  &  \textbf{89.59$\pm$0.23}     & \textbf{89.48$\pm$0.17}  &  84.13$\pm$1.30\\ \hline

\multirow{4}{*}{REUT}  & ACC    & 54.04$\pm$0.01 & 74.90$\pm$0.21 & 73.58$\pm$0.13 & 75.43$\pm$0.14 & 54.40$\pm$0.27 & 60.85$\pm$0.23 & 65.50$\pm$0.13 & \textit{79.30$\pm$0.11} & \rm{79.30$\pm$1.07}  &  \textbf{80.32$\pm$0.04}     &  \underline{79.35$\pm$0.00}  &  \textbf{81.78$\pm$0.01}\\
                          & NMI    & 41.54$\pm$0.51 & 49.69$\pm$0.29 & 47.50$\pm$0.34 & 50.28$\pm$0.17 & 25.92$\pm$0.41 & 25.51$\pm$0.22 & 30.55$\pm$0.29 & \textit{56.89$\pm$0.27} & \underline{57.83$\pm$1.01}  &  55.63$\pm$0.05      &  55.16$\pm$0.01  & \textbf{59.13$\pm$0.00}  \\
                          & ARI    & 27.95$\pm$0.38 & 49.55$\pm$0.37 & 48.44$\pm$0.14 & 51.26$\pm$0.21 & 19.61$\pm$0.22 & 26.18$\pm$0.36 & 31.12$\pm$0.18 & \textit{59.58$\pm$0.32} & \underline{60.55$\pm$1.78}  &  \underline{59.67$\pm$0.11}      &  57.80$\pm$0.01  & \textbf{63.51$\pm$0.03} \\
                          & F1     & 41.28$\pm$2.43 & 60.96$\pm$0.22 & 64.25$\pm$0.22 & 63.21$\pm$0.12 & 43.53$\pm$0.42 & 57.14$\pm$0.17 & 61.82$\pm$0.13 & \textit{66.15$\pm$0.15} & \rm{66.16$\pm$0.64}  &  63.66$\pm$0.03      &  \underline{66.54$\pm$0.01} &  \textbf{69.48$\pm$0.03}\\ \hline
\end{tabular}
}
\end{table*}

\subsection{Quantitative Results}
We show our hyperparameters in Table~\ref{TABLE_datasets}. Comparison of results with state-of-the-art graph and non-graph datasets are in Table~\ref{table_graph_results} and Table~\ref{table_non_graph_results}, respectively. For the graph data, \acrshort{OURMODEL} is state-of-the-art for DBLP. A paired-t test shows ACM and CITE results to be best for both SCGC* and \acrshort{OURMODEL}.  In non-graph results, \acrshort{OURMODEL} comes second best in USPS image data. While results for HHAR are somewhat lagging, \acrshort{OURMODEL} is the best for REUT. Generally we achieve better results on the natural graph datasets; ACM, DBLP and CITE, while being competitive on other modalities. 

\subsection{Performance}
\begin{figure}
  \centering
  \includegraphics[width=\textwidth]{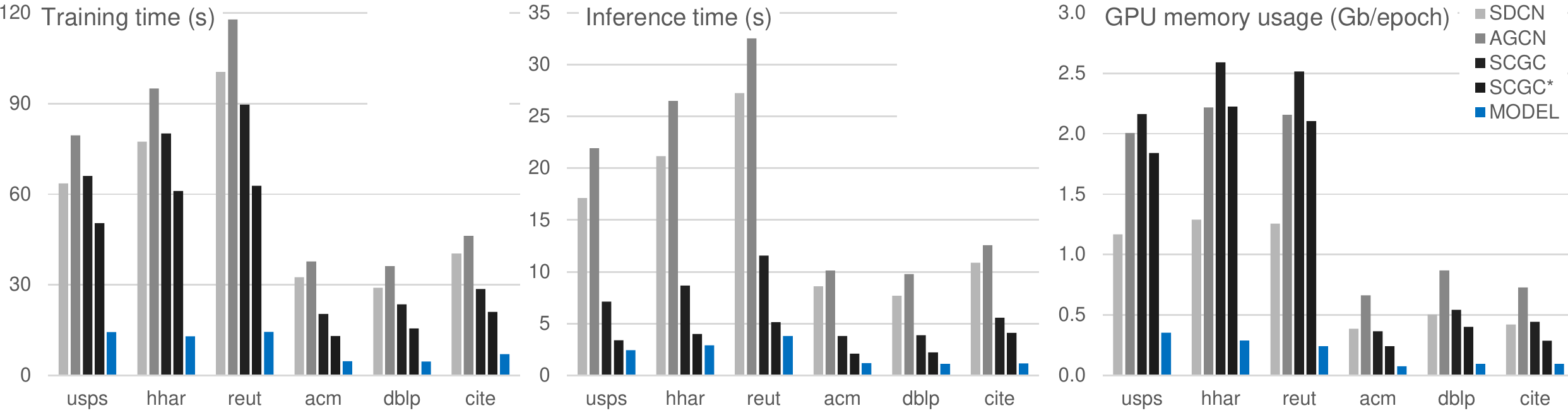}
  \caption{GPU performance from the pytorch profiler on Google Colab with T4 16Gb GPU. \textbf{left}:training time for 200 epochs. \textbf{centre}:inference for 200 runs. \textbf{right}:memory utilization per epoch.}
  \label{fig_training_memory}
\end{figure}

In Figure~\ref{fig_training_memory} we compare the GPU based training, inference times and GPU memory. Our model times also include the time taken for the cumulative influence computation. For all the datasets, \acrshort{OURMODEL} is superior by 3.6x training time, 1.8x inference time and 5.3x GPU memory savings. Especially, for larger datasets USPS, HHAR and REUT, \acrshort{OURMODEL} uses 5.2,7.7,8.7x less GPU memory. Note that SCGC only differs from \acrshort{OURMODEL} by its use of the novel proxy-ed \acrshort{LOSS} to which we solely attribute the time and memory savings. 

\subsection{Qualitative Results}
\begin{figure}[h]
  \centering
  \includegraphics[width=0.85\columnwidth]{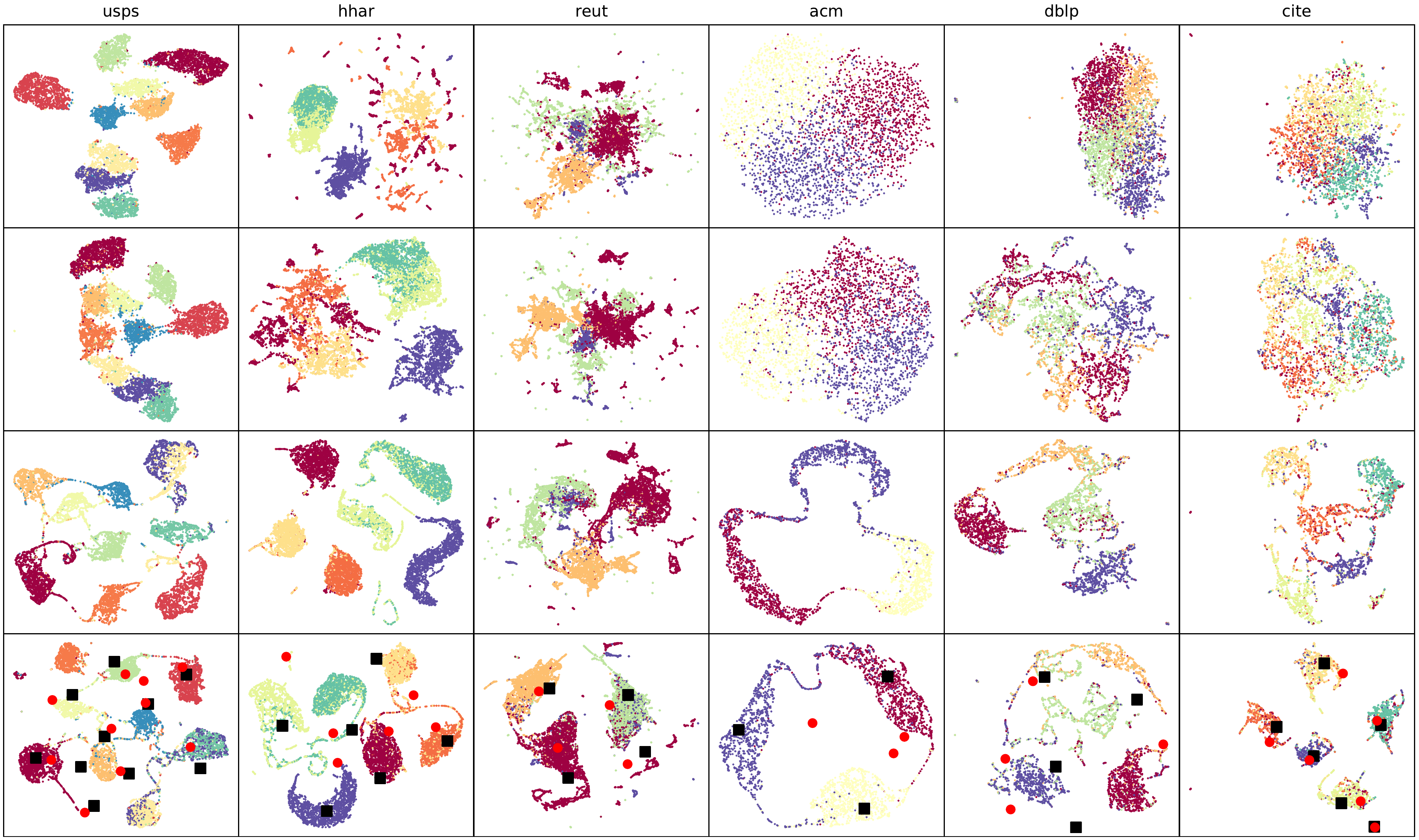}
  \caption{Visual comparison of embeddings; top: raw data, second row: after AE pre-training, third-row: from SCGC*, and last-row: from \acrshort{OURMODEL}*. Colors represent ground truth groups. Black squares, $\hat{\mu}$, are the approximated meta-nodes. Red dots, $\mu$, are the cluster centroids.}
  \label{fig_embeddings}
\end{figure}
We use UMAP \citep{mcinnes2018umap}, in Figure~\ref{fig_embeddings}, to get a visual understanding of the raw and learnt embedding spaces. Except for USPS, which is a distinct set of $0 \cdots 9$ handwritten digits (raw 1), we see that all other datasets produce quite indistinguishable clusters. Clustering is nearly non-existent in the (last 3) graph datasets. This clearly shows a characteristic difference in graph data, which can lead to high samplings bias. Note that $\hat{\mu} \ne \mu$ for any meta-node.

\subsection{Ablation study}
\begin{figure}[h]
  \centering
  \includegraphics[width=0.8\columnwidth]{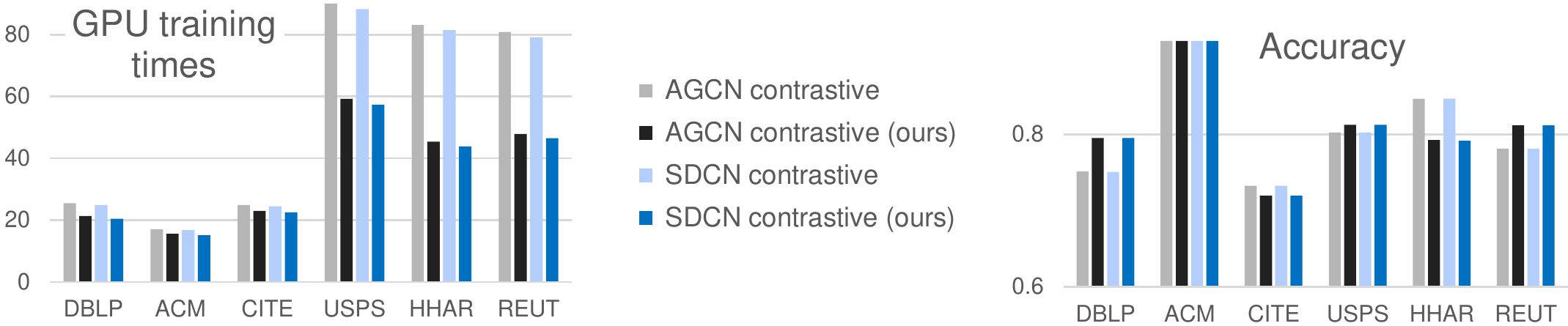}
  \caption{Left:GPU training times with \acrshort{OURMODEL} on SDCN and AGCN is consistently lower, and significant in large datasets (usps,hhar,reut). Right:Accuracy loss of the approximation is very low. For dblp, usps, reut accuracy is actually better.}
  \label{fig_ablation_GPU_Acc}
\end{figure}
To investigate \acrshort{LOSS}s ability to generalize to other models, we incorporate it to SDCN and AGCN models, modified for contrastive loss. 
Figure~\ref{fig_ablation_GPU_Acc} shows the GPU training time and accuracy. 
As \acrshort{LOSS} is a loss function, there is no difference in the inference times. As expected, training times are significantly shorter, with (often better) training accuracy due to block contrastiveness. Note that \acrshort{LOSS} only improves loss computation efficiency. Majority of the SDCN and AGCN computation time is spent in their GNNs convolution operations.  

We also carry out extensive experimentation to assess the behavior of hyperparameters. \acrshort{LOSS} is robust to changes in hyperparameter values and performs best with a learning rate of $0.001$, as shown in Appendix~\ref{Appendix_CCONLYHYPER}. Further, \acrshort{LOSS} accuracy against all hyperparameter combinations is generally equal or better than the less efficient non proxy-ed contrastive loss variant, as seen in Appendix~\ref{Appendix_NNCCHYPER}.

\subsection{Future work}
Our parameter-free proxy-ed contrastive loss uses the full positive edge information which, as some of our experiments has shown, is redundant. For example, USPS gives similar results with 40\% positive edges removed. An algorithm to drop un-informative edges may result in further efficiency improvements, which we leave for future work.
While theoretically possible, it would be interesting to see how our proxy-ed contrastive loss works with semi or fully supervised data. Further study is needed to explore how hard cluster centers effect the optimization process.
Contrastive loss for vision typically contrasts $N$ anchor images, each augmented to create two views of the same sample $x_i, x_j$ with the remaining $2(N-1)$ \citep{chen2020simpleVisual, chuang2020debiased}. While this approach is different from ours, we have shown a possible use case with USPS for vision data.

\section{Conclusion}\label{Conclusion}
In this work, we present an efficient parameter-free proxy approximation to incorporate negative samples in contrastive loss for joint clustering and representation learning. We eliminate sample bias, achieve block contrastiveness and $0(N)$. Our work is supported by theoretical proof and empirical results.
We improve considerably over previous methods accuracy, speed and memory usage. 
Our approach differs from prior self-supervised clustering by the proxy mechanism we use to incorporate \textbf{all} negative samples efficiently. The strength of this simple approach indicates that, despite the increased interest in graphs, effective contrastive learning remains relatively unexplored.

\section*{Acknowledgments}
Dedicated to Sugandi.
\clearpage
\bibliography{main}

\begin{thebibliography}{29}
\providecommand{\natexlab}[1]{#1}
\providecommand{\url}[1]{\texttt{#1}}
\expandafter\ifx\csname urlstyle\endcsname\relax
  \providecommand{\doi}[1]{doi: #1}\else
  \providecommand{\doi}{doi: \begingroup \urlstyle{rm}\Url}\fi

\bibitem[Altman(1992)]{altman1992introduction}
Naomi~S Altman.
\newblock An introduction to kernel and nearest-neighbor nonparametric
  regression.
\newblock \emph{The American Statistician}, 46\penalty0 (3):\penalty0 175--185,
  1992.

\bibitem[Arora et~al.(2019)Arora, Khandeparkar, Khodak, Plevrakis, and
  Saunshi]{arora2019theoretical}
Sanjeev Arora, Hrishikesh Khandeparkar, Mikhail Khodak, Orestis Plevrakis, and
  Nikunj Saunshi.
\newblock A theoretical analysis of contrastive unsupervised representation
  learning.
\newblock In \emph{36th International Conference on Machine Learning, ICML
  2019}, pages 9904--9923. International Machine Learning Society (IMLS), 2019.

\bibitem[Bo et~al.(2020)Bo, Wang, Shi, Zhu, Lu, and Cui]{bo2020structural}
Deyu Bo, Xiao Wang, Chuan Shi, Meiqi Zhu, Emiao Lu, and Peng Cui.
\newblock Structural deep clustering network.
\newblock In \emph{Proceedings of The Web Conference 2020}, pages 1400--1410,
  2020.

\bibitem[Chen et~al.(2020{\natexlab{a}})Chen, Wei, Huang, Ding, and
  Li]{chen2020simple}
Ming Chen, Zhewei Wei, Zengfeng Huang, Bolin Ding, and Yaliang Li.
\newblock Simple and deep graph convolutional networks.
\newblock In \emph{International Conference on Machine Learning}, pages
  1725--1735. PMLR, 2020{\natexlab{a}}.

\bibitem[Chen et~al.(2020{\natexlab{b}})Chen, Kornblith, Norouzi, and
  Hinton]{chen2020simpleVisual}
Ting Chen, Simon Kornblith, Mohammad Norouzi, and Geoffrey Hinton.
\newblock A simple framework for contrastive learning of visual
  representations.
\newblock In \emph{International conference on machine learning}, pages
  1597--1607. PMLR, 2020{\natexlab{b}}.

\bibitem[Chuang et~al.(2020)Chuang, Robinson, Lin, Torralba, and
  Jegelka]{chuang2020debiased}
Ching-Yao Chuang, Joshua Robinson, Yen-Chen Lin, Antonio Torralba, and Stefanie
  Jegelka.
\newblock Debiased contrastive learning.
\newblock \emph{Advances in neural information processing systems},
  33:\penalty0 8765--8775, 2020.

\bibitem[Golub and Reinsch(1971)]{golub1971singular}
Gene~H Golub and Christian Reinsch.
\newblock Singular value decomposition and least squares solutions.
\newblock In \emph{Linear algebra}, pages 134--151. Springer, 1971.

\bibitem[Guo et~al.(2017)Guo, Gao, Liu, and Yin]{guo2017improved}
Xifeng Guo, Long Gao, Xinwang Liu, and Jianping Yin.
\newblock Improved deep embedded clustering with local structure preservation.
\newblock In \emph{IJCAI}, pages 1753--1759, 2017.

\bibitem[Hamilton et~al.(2017)Hamilton, Ying, and
  Leskovec]{hamilton2017inductive}
William~L Hamilton, Rex Ying, and Jure Leskovec.
\newblock Inductive representation learning on large graphs.
\newblock In \emph{Proceedings of the 31st International Conference on Neural
  Information Processing Systems}, pages 1025--1035, 2017.

\bibitem[Hartigan and Wong(1979)]{hartigan1979algorithm}
John~A Hartigan and Manchek~A Wong.
\newblock Algorithm as 136: A k-means clustering algorithm.
\newblock \emph{Journal of the Royal Statistical Society. Series C (Applied
  Statistics)}, 28\penalty0 (1):\penalty0 100--108, 1979.

\bibitem[Hinton and Salakhutdinov(2006)]{hinton2006reducing}
Geoffrey~E Hinton and Ruslan~R Salakhutdinov.
\newblock Reducing the dimensionality of data with neural networks.
\newblock \emph{Science}, 313\penalty0 (5786):\penalty0 504--507, 2006.

\bibitem[Hu et~al.(2021)Hu, You, Wang, Wang, Zhou, and Gao]{hu2021graph}
Yang Hu, Haoxuan You, Zhecan Wang, Zhicheng Wang, Erjin Zhou, and Yue Gao.
\newblock Graph-mlp: node classification without message passing in graph.
\newblock \emph{arXiv preprint arXiv:2106.04051}, 2021.

\bibitem[Kipf and Welling(2016)]{kipf2016variational}
Thomas~N Kipf and Max Welling.
\newblock Variational graph auto-encoders.
\newblock \emph{arXiv preprint arXiv:1611.07308}, 2016.

\bibitem[Kulatilleke et~al.(2021)Kulatilleke, Portmann, Ko, and
  Chandra]{kulatilleke2021fdgatii}
Gayan~K Kulatilleke, Marius Portmann, Ryan Ko, and Shekhar~S Chandra.
\newblock Fdgatii: Fast dynamic graph attention with initial residual and
  identity mapping.
\newblock \emph{arXiv preprint arXiv:2110.11464}, 2021.

\bibitem[Kulatilleke et~al.(2022)Kulatilleke, Portmann, and
  Chandra]{kulatilleke2022scgc}
Gayan~K Kulatilleke, Marius Portmann, and Shekhar~S Chandra.
\newblock Scgc: Self-supervised contrastive graph clustering.
\newblock \emph{arXiv preprint arXiv:2204.12656}, 2022.

\bibitem[Le~Cun et~al.(1990)Le~Cun, Matan, Boser, Denker, Henderson, Howard,
  Hubbard, Jacket, and Baird]{le1990handwritten}
Yann Le~Cun, Ofer Matan, Bernhard Boser, John~S Denker, Don Henderson,
  Richard~E Howard, Wayne Hubbard, LD~Jacket, and Henry~S Baird.
\newblock Handwritten zip code recognition with multilayer networks.
\newblock In \emph{ICPR}, volume~2, pages 35--40. IEEE, 1990.

\bibitem[Lewis et~al.(2004)Lewis, Yang, Rose, and Li]{lewis2004rcv1}
David~D Lewis, Yiming Yang, Tony~G Rose, and Fan Li.
\newblock Rcv1: A new benchmark collection for text categorization research.
\newblock \emph{Journal of machine learning research}, 5\penalty0
  (Apr):\penalty0 361--397, 2004.

\bibitem[Logeswaran and Lee(2018)]{logeswaran2018efficient}
Lajanugen Logeswaran and Honglak Lee.
\newblock An efficient framework for learning sentence representations.
\newblock In \emph{International Conference on Learning Representations}, 2018.

\bibitem[Maaten and Hinton(2008)]{maaten2008visualizing}
Laurens van~der Maaten and Geoffrey Hinton.
\newblock Visualizing data using t-sne.
\newblock \emph{Journal of machine learning research}, 9\penalty0
  (Nov):\penalty0 2579--2605, 2008.

\bibitem[McInnes et~al.(2018)McInnes, Healy, and Melville]{mcinnes2018umap}
Leland McInnes, John Healy, and James Melville.
\newblock Umap: Uniform manifold approximation and projection for dimension
  reduction.
\newblock \emph{arXiv preprint arXiv:1802.03426}, 2018.

\bibitem[Pan et~al.(2018)Pan, Hu, Long, Jiang, Yao, and
  Zhang]{pan2018adversarially}
Shirui Pan, Ruiqi Hu, Guodong Long, Jing Jiang, Lina Yao, and Chengqi Zhang.
\newblock Adversarially regularized graph autoencoder for graph embedding.
\newblock \emph{arXiv preprint arXiv:1802.04407}, 2018.

\bibitem[Park et~al.(2022)Park, Rossi, Koh, Burhanuddin, Kim, Du, Ahmed, and
  Faloutsos]{park2022cgc}
Namyong Park, Ryan Rossi, Eunyee Koh, Iftikhar~Ahamath Burhanuddin, Sungchul
  Kim, Fan Du, Nesreen Ahmed, and Christos Faloutsos.
\newblock Cgc: Contrastive graph clustering forcommunity detection and
  tracking.
\newblock In \emph{Proceedings of the ACM Web Conference 2022}, pages
  1115--1126, 2022.

\bibitem[Peng et~al.(2021)Peng, Liu, Jia, and Hou]{peng2021attention}
Zhihao Peng, Hui Liu, Yuheng Jia, and Junhui Hou.
\newblock Attention-driven graph clustering network.
\newblock In \emph{Proceedings of the 29th ACM International Conference on
  Multimedia}, pages 935--943, 2021.

\bibitem[Stisen et~al.(2015)Stisen, Blunck, Bhattacharya, Prentow,
  Kj{\ae}rgaard, Dey, Sonne, and Jensen]{stisen2015smart}
Allan Stisen, Henrik Blunck, Sourav Bhattacharya, Thor~Siiger Prentow,
  Mikkel~Baun Kj{\ae}rgaard, Anind Dey, Tobias Sonne, and Mads~M{\o}ller
  Jensen.
\newblock Smart devices are different: Assessing and mitigatingmobile sensing
  heterogeneities for activity recognition.
\newblock In \emph{SenSys}, pages 127--140, 2015.

\bibitem[Velickovic et~al.(2019)Velickovic, Fedus, Hamilton, Li{\`o}, Bengio,
  and Hjelm]{velickovic2019deep}
Petar Velickovic, William Fedus, William~L Hamilton, Pietro Li{\`o}, Yoshua
  Bengio, and R~Devon Hjelm.
\newblock Deep graph infomax.
\newblock \emph{ICLR (Poster)}, 2\penalty0 (3):\penalty0 4, 2019.

\bibitem[Wang et~al.(2019)Wang, Pan, Hu, Long, Jiang, and
  Zhang]{wang2019attributed}
Chun Wang, Shirui Pan, Ruiqi Hu, Guodong Long, Jing Jiang, and Chengqi Zhang.
\newblock Attributed graph clustering: A deep attentional embedding approach.
\newblock \emph{arXiv preprint arXiv:1906.06532}, 2019.

\bibitem[Wang and Isola(2020)]{wang2020understanding}
Tongzhou Wang and Phillip Isola.
\newblock Understanding contrastive representation learning through alignment
  and uniformity on the hypersphere.
\newblock In \emph{International Conference on Machine Learning}, pages
  9929--9939. PMLR, 2020.

\bibitem[Xie et~al.(2016)Xie, Girshick, and Farhadi]{xie2016unsupervised}
Junyuan Xie, Ross Girshick, and Ali Farhadi.
\newblock Unsupervised deep embedding for clustering analysis.
\newblock In \emph{ICML}, pages 478--487, 2016.

\bibitem[Zhu et~al.(2021)Zhu, Sun, and Koniusz]{zhu2021contrastive}
Hao Zhu, Ke~Sun, and Peter Koniusz.
\newblock Contrastive laplacian eigenmaps.
\newblock \emph{Advances in Neural Information Processing Systems}, 34, 2021.

\end{thebibliography}
\bibliographystyle{plainnat}

\clearpage
\appendix
\section{Appendix}
\subsection{Proofs of Theoretical Results - Derivation of Equation~\ref{EQ_INE}} \label{Appendix_PROOF1}
Assume node embeddings $Z=\{z_1,z_2,z_3 \dots z_N\}$, clusters $\mu=\{\mu_1,\mu_2 \dots \mu_C\}$, a label assignment operator $\operatorname{label}(z_i)$ such that $\mu_a = \sum_{i=1}^{N} \mathbf{1}_{[i \in \operatorname{label}(z_i)=a]}\cdot z_i$, a hyperparameter $\tau$ related to the temperature in contrastive loss and 
\begin{equation}
     \operatorname{similarity}(i, j, z_i,z_j) = \operatorname{sim}(z_i,z_j) \begin{cases}
        0,& \text{$i=j$} \\
         \frac{z_i \cdot z_j}{\|z_i\| \|z_j\|} ,& \text{$i \ne j$}\\
\end{cases}
\label{EQA_SIM}
\end{equation}
We use $\operatorname{sim}(z_i,z_j)$ as the shorthand notation for $\operatorname{similarity}(i, j, z_i,z_j)$ interchangeably for brevity.

We begin with Equation~\ref{EQ_TRADITIONAL}, which is the popular form of contrastive loss \citep{hu2021graph,kulatilleke2022scgc}. With $\tau$ as the temperature parameter, $\gamma_{ij}$ the relationship between nodes $i,j$, the loss for the $i^{th}$ can be expanded as: 
\begin{equation}
\ell_{i}=
+\log {\sum_{j=1}^{B} \mathbf{1}_{[j \neq i]} \exp \left(\operatorname{sim}\left(\boldsymbol{z}_{i}, \boldsymbol{z}_{j}\right) \tau\right)}
-\log {\sum_{j=1}^{B} \mathbf{1}_{[j \neq i]}   \gamma_{ij}  \exp \left(\operatorname{sim}\left(\boldsymbol{z}_{i}, \boldsymbol{z}_{j}\right) \tau\right)},
\label{EQ_TRADITIONAL2}
\end{equation}

where, the first part on the right corresponds to the negative node contrasting portion and the second portion contrasts the positives for node $i$. 
From Equation~\ref{EQ_TRADITIONAL2}, for all nodes $N$, we take to negative node contrasting portion, by averaging over $N$ nodes to obtain: 
\begin{equation}
loss_{NN} = \frac{1}{N}  \sum _{i=1}^{N} log\left[ 
    \sum _{j=1}^{N} e^{ \operatorname{sim}(i,j,z_i,z_j)\tau}
\right],
\label{EQA_LOSSNN}
\end{equation}
Note our use of the more concise $\operatorname{sim()}$ and the compact $e$ notation over $\exp()$ interchangeably for compactness reasons.  

We expand Equation~\ref{EQA_LOSSNN}, together with $e^0=1$ in cases where $i=j$, as:
\begin{align}
  loss_{NN} =  \frac{1}{N} \Bigg[ \Bigg.  & log\left( \quad \quad 1 \quad \quad + e^{sim(z_1,z_2)\tau} + e^{sim(z_1,z_3)\tau} + e^{sim(z_1,z_4)\tau} \ldots + e^{sim(z_1,z_N)\tau} \right) +  \nonumber\\
											  & log\left( e^{sim(z_2,z_1)\tau} + \quad \quad 1 \quad \quad + e^{sim(z_2,z_3)\tau} + e^{sim(z_2,z_4)\tau} \ldots + e^{sim(z_2,z_N)\tau} \right) +  \nonumber\\
											  & log\left( e^{sim(z_3,z_1)\tau} + e^{sim(z_3,z_2)\tau} + \quad \quad 1 \quad \quad + e^{sim(z_3,z_4)\tau} \ldots + e^{sim(z_3,z_N)\tau} \right) +  \nonumber\\
											  & \quad \quad \quad \quad \quad \quad \quad \quad \quad \quad \quad \quad \cdots  \nonumber\\
											  & log\left( e^{sim(z_N,z_1)\tau} + e^{sim(z_N,z_2)\tau} + e^{sim(z_N,z_3)\tau} + e^{sim(z_N,z_4)\tau} \ldots + \quad 1 \right)    \Bigg. \Bigg]
\label{EQA_LOSSNNlong}											  
\end{align}

Similarly, we can express the cluster based contrastive loss as:
\begin{equation}
loss_{CC} = \frac{1}{C}  \sum _{a=1}^{C} log\left[ 
    \sum _{b=1}^{M} e^{ \operatorname{sim}(a,b,\mu_a,\mu_b)\tau}
\right]
\label{EQA_LOSSCC}
\end{equation}
with the following expansion:
\begin{align}
  loss_{CC}=  \frac{1}{C} \Bigg[ \Bigg.  & log\left( \quad \quad 1 \quad \quad + e^{sim(\mu_1,\mu_2)\tau} + e^{sim(\mu_1,\mu_3)\tau} + e^{sim(\mu_1,\mu_4)\tau} \ldots + e^{sim(\mu_1,\mu_C)\tau} \right) +  \nonumber\\
											  & log\left( e^{sim(\mu_2,\mu_1)\tau} + \quad \quad 1 \quad \quad + e^{sim(\mu_2,\mu_3)\tau} + e^{sim(\mu_2,\mu_4)\tau} \ldots + e^{sim(\mu_2,\mu_C)\tau} \right) +  \nonumber\\
											  & log\left( e^{sim(\mu_3,\mu_1)\tau} + e^{sim(\mu_3,\mu_2)\tau} + \quad \quad 1 \quad \quad + e^{sim(\mu_3,\mu_4)\tau} \ldots + e^{sim(\mu_3,\mu_C)\tau} \right) +  \nonumber\\
											  & \quad \quad \quad \quad \quad \quad \quad \quad \quad \quad \quad \quad \cdots  \nonumber\\
											  & log\left( e^{sim(\mu_C,\mu_1)\tau} + e^{sim(\mu_C,\mu_2)\tau} + e^{sim(\mu_C,\mu_3)\tau} + e^{sim(\mu_C,\mu_4)\tau} \ldots + \quad 1 \right)  \Bigg. \Bigg]
\label{EQA_LOSSCClong}											  
\end{align}

If, $loss_{NN}^{min} > loss_{CC}^{max}$, we have $\frac{loss_{NN}}{loss_{CC}} > 1$. Next we show the conditions necessary for establishing this inequality.

As $ 0 \leq \operatorname{sim} \leq 1.0$, we obtain the $min$ using $\operatorname{sim}_{min} = 0$:
\begin{align}
  loss_{NN}^{min} &=  \frac{1}{N} \Bigg[ \Bigg.  log\left(1 + e^0 + e^0 + \ldots + e^0 \right) + \cdots + log\left(1 + e^0 + e^0 + \ldots + e^0 \right) \Bigg] \nonumber\\
  &= log \left[ 1 + (N-1)e^0 \right] \nonumber\\
  &= log(N)
\label{EQA_LOSSNNmin}											  
\end{align}

Similarly, we can obtain the $max$, using $\operatorname{sim}_{max} = 1.0$:
\begin{align}
  loss_{CC}^{max} &=  \frac{1}{C} \Bigg[ \Bigg.  log\left(1 + e^{1.\tau} + e^{1.\tau} + \ldots + e^{1.\tau} \right) + \cdots + log\left(1 + e^{1.\tau} + e^{1.\tau} + \ldots + e^{1.\tau} \right) \Bigg] \nonumber\\
  &= log \left[ 1 + (C-1)e^{\tau}\right]
\label{EQA_LOSSCCmax}											  
\end{align}

Combining Equation~\ref{EQA_LOSSNNmin} and Equation~\ref{EQA_LOSSCCmax}, we establish the necessary condition for our inequality, Equation~\ref{EQ_INE} as;
\begin{align*}
  \frac{loss_{NN}}{loss_{CC}} >  \frac{log(N)}{log \left[ 1 + (C-1)e^{\tau}\right]}
\label{EQA_INE}											  
\end{align*}
This derivation is used in Section~\ref{Approximation}, where we show how the condition is almost always satisfied in real graphs. As a result, $loss_{NN}$ upper bounds $loss_{CC}$. Note that a lower loss is better.

\clearpage
\subsection{Hyperparameters vs Accuracy} \label{Appendix_CCONLYHYPER}
\begin{figure}[h]
  \centering
  \includegraphics[width=0.9\columnwidth]{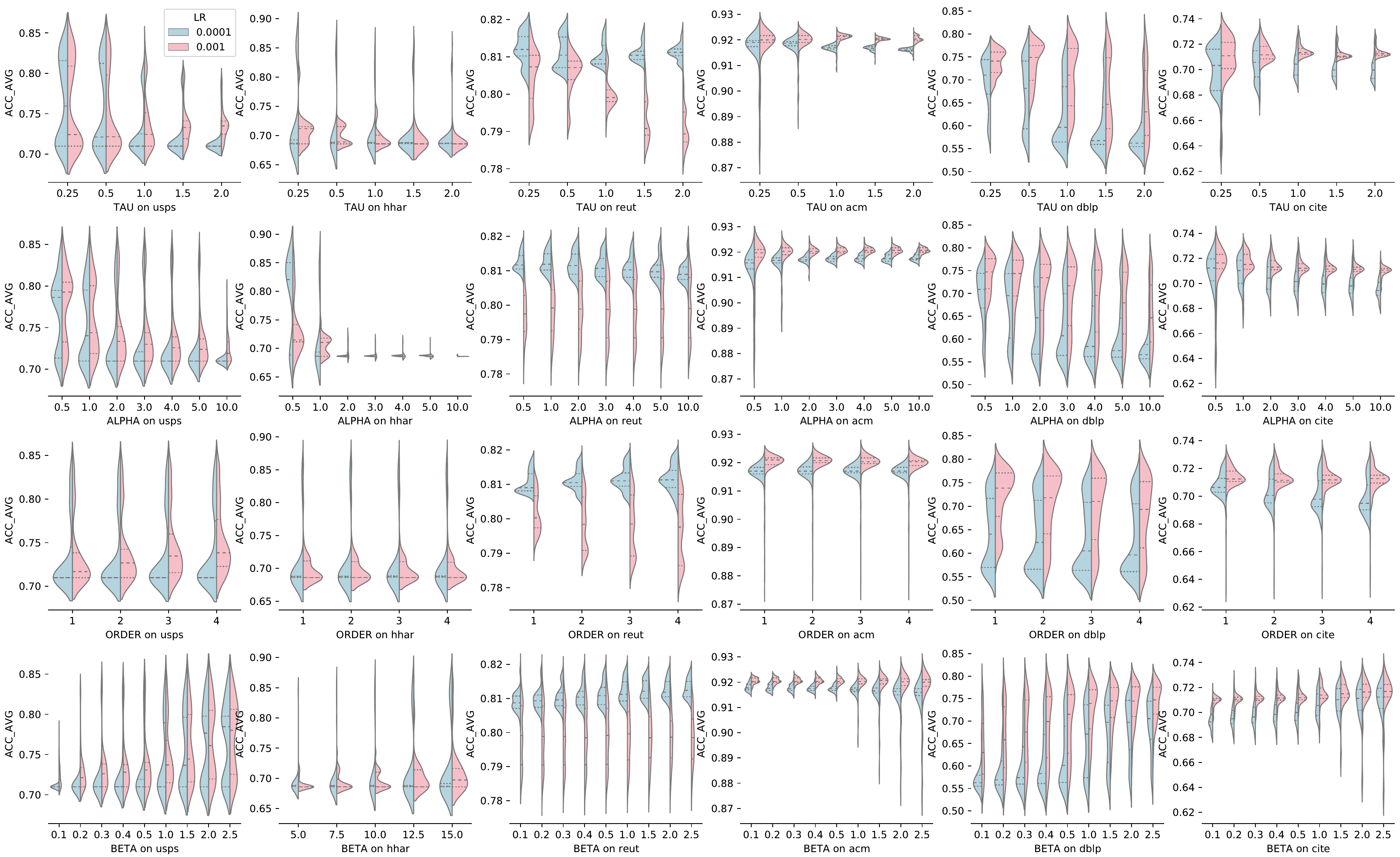}
  \caption{Ablation study on the hyperparameters. TAU=$\tau$, ALPHA=$\alpha$, ORDER=$R$ and LR denotes learning rate. A hyperparameter with higher and more condensed distribution represents its superiority over its counterpart. 
  \acrshort{OURMODEL} is robust to $\tau, \alpha, R$ and best with a learning rate 0f $0.001$.
  Best viewed in colour.}
  \label{fig_hyper_CCOnly}
\end{figure}

\subsection{Hyperparameter behaviour with and without \acrshort{LOSS}} \label{Appendix_NNCCHYPER}
\begin{figure}[h]
  \centering
  \includegraphics[width=0.9\columnwidth]{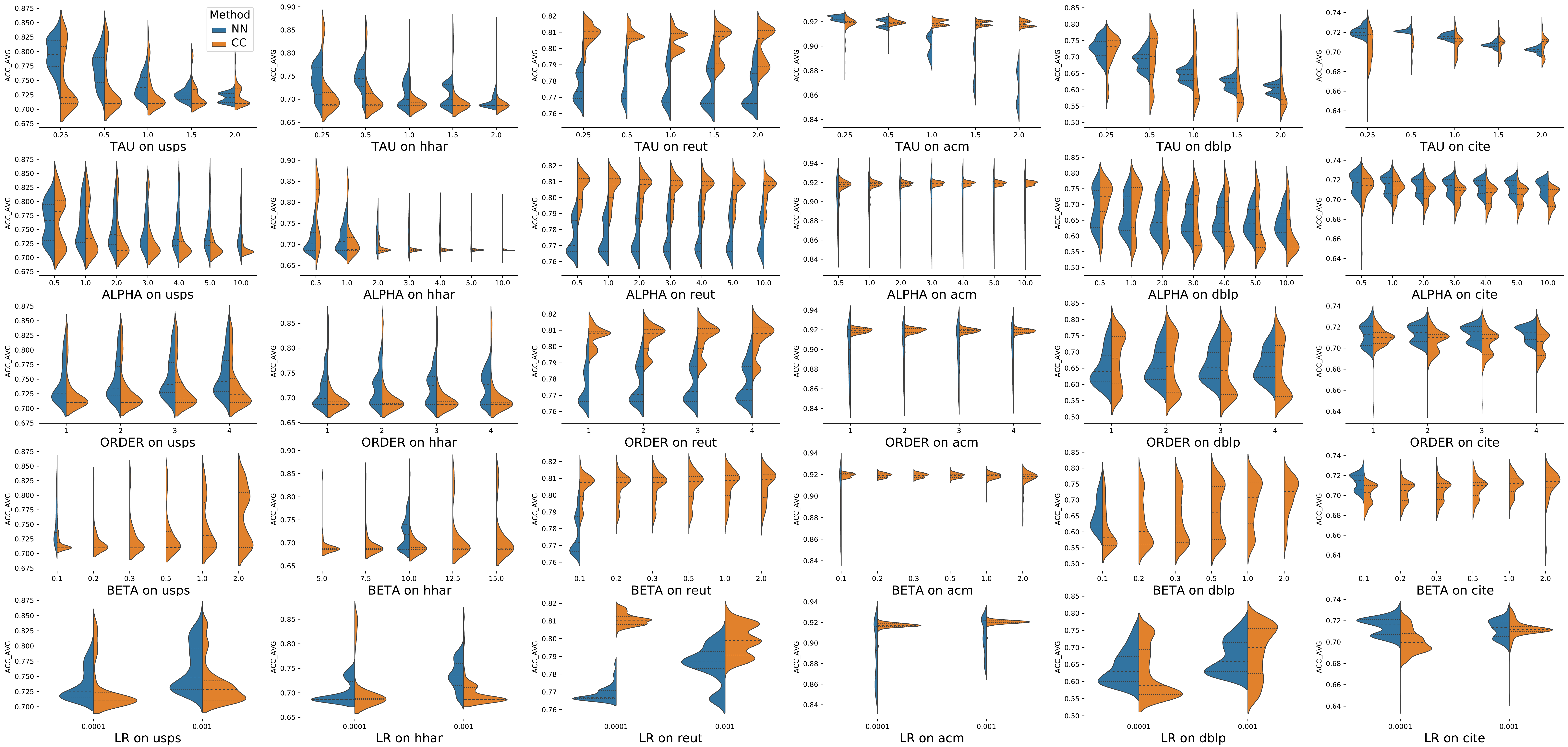}
  \caption{Comparison of hyperparameters with and without \acrshort{LOSS}. TAU=$\tau$, ALPHA=$\alpha$, ORDER=$R$ and LR denotes learning rate. A hyperparameter with higher and more condensed distribution represents its superiority over its counterpart. 
  \acrshort{LOSS} is generally better in accuracy for majority of the hyperparameter combinations.
  Best viewed in colour.}
  \label{fig_hyper_NNCC}
\end{figure}

\end{document}